\documentclass[10pt,twocolumn,letterpaper]{article}

\usepackage{cvpr}              
\usepackage{tabularx}
\definecolor{cvprblue}{rgb}{0.21,0.49,0.74}
\usepackage[pagebackref,breaklinks,colorlinks,allcolors=cvprblue]{hyperref}
\usepackage{multirow}
\usepackage{algorithm} 
\usepackage{algorithmic}
\usepackage{makecell}
\def\paperID{9097} 
\def\confName{CVPR}
\def\confYear{2026}

\title{TrajDiff: End-to-end Autonomous Driving without Perception Annotation}

\author{Xingtai Gui$^1$,  Jianbo Zhao$^2$,  Wencheng Han$^1$, Jikai Wang$^1$, \\
Jiahao Gong$^3$, Feiyang Tan$^3$,  Cheng-zhong Xu$^1$, Jianbing Shen$^{1*}$ \\
$^{1}$SKL-IOTSC, CIS, University of Macau,
$^{2}$University of Science and Technology of China, \\
$^{3}$Mach Drive
}

\begin{document}
\maketitle
\begin{abstract}
End-to-end autonomous driving systems directly generate driving policies from raw sensor inputs. While these systems can extract effective environmental features for planning, relying on auxiliary perception tasks, developing perception annotation-free planning paradigms has become increasingly critical due to the high cost of manual perception annotation. In this work, we propose TrajDiff, a Trajectory-oriented BEV Conditioned Diffusion framework that establishes a fully perception annotation-free generative method for end-to-end autonomous driving. TrajDiff requires only raw sensor inputs and future trajectory, constructing Gaussian BEV heatmap targets that inherently capture driving modalities. We design a simple yet effective trajectory-oriented BEV encoder to extract the TrajBEV feature without perceptual supervision. Furthermore, we introduce Trajectory-oriented BEV Diffusion Transformer (TB-DiT), which leverages ego-state information and the predicted TrajBEV features to directly generate diverse yet plausible trajectories, eliminating the need for handcrafted motion priors. Beyond architectural innovations, TrajDiff enables exploration of data scaling benefits in the annotation-free setting. Evaluated on the NAVSIM benchmark, TrajDiff achieves 87.5 PDMS, establishing state-of-the-art performance among all annotation-free methods. With data scaling, it further improves to 88.5 PDMS, which is comparable to advanced perception-based approaches. Our code and model will be made publicly available.

\end{abstract}    
\section{Introduction}
\label{sec:intro}

End-to-end autonomous driving has gained significant attention in recent research, distinguished by its capability to directly derive driving policies from raw sensor inputs. While individual modules such as perception~\cite{li2024bevformer, huang2021bevdet, liu2022petr}, prediction~\cite{tang2024hpnet, feng2024unitraj, gui2024fiptr}, and planning~\cite{cheng2024pluto, scheel2022urban, cheng2024rethinking} have achieved remarkable progress through advanced model design, end-to-end methods not only incorporate these module-level advancements but also fundamentally shift the optimization focus to direct planning performance~\cite{hu2023planning, jiang2023vad, shao2023reasonnet}. The data-driven nature of the end-to-end framework obviates the need for heuristic-based modular interaction, thereby mitigating information loss and cascading failures in conventional pipelines.

\begin{figure}[t]
  \centering
  \includegraphics[width=1\linewidth]{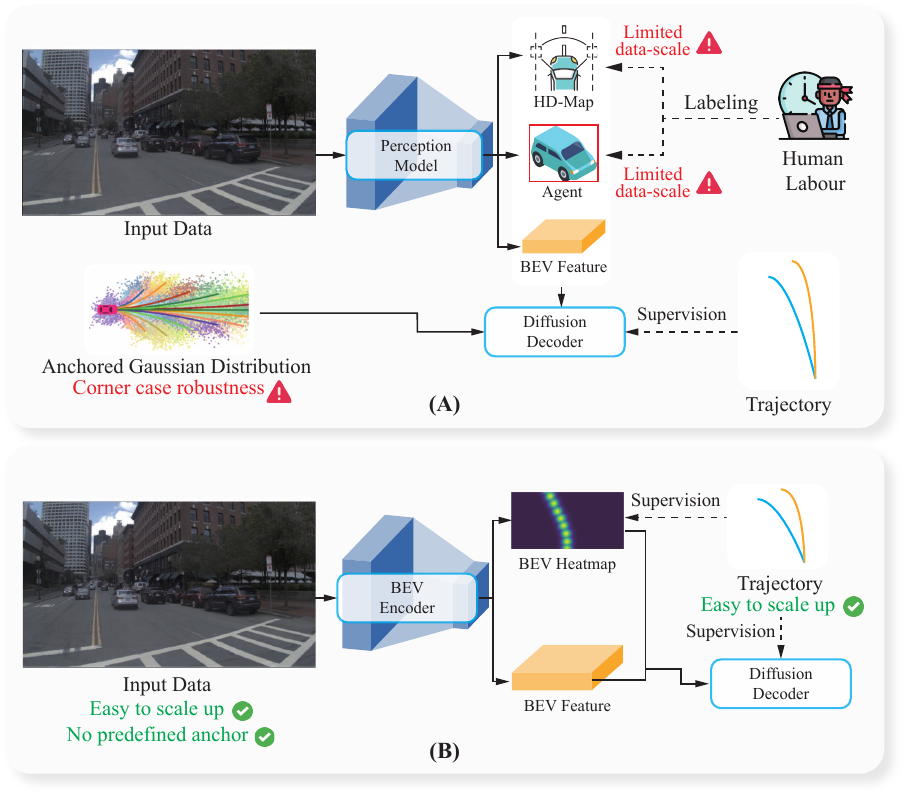}
  \caption{\textbf{The comparison of different end-to-end autonomous driving paradigms.} (A)~Perception annotation-based and anchor-based diffusion framework. (B)~Our proposed TrajDiff leverages a BEV heatmap and trajectory-oriented BEV feature, effectively realizing perception annotation-free and anchor-free end-to-end autonomous driving, which is easy to benefit from data scaling up.}
\label{Fig:motivation}
\end{figure}

To address the inherent uncertainty and multi-modality of motion planning, generative end-to-end approaches are emerging as a mainstream paradigm~\cite{chen2024vadv2, li2024hydra, li2025hydra}. Inspired by the remarkable generative capabilities of diffusion models in image generation~\cite{ho2020denoising, peebles2023scalable}, video generation~\cite{blattmann2023stable, yang2024cogvideox}, and trajectory prediction \cite{jiang2023motiondiffuser, wang2024optimizing, bae2024singulartrajectory}, recent advances have successfully adapted diffusion policy for end-to-end driving. Their principle is to directly sample trajectories from noisy trajectories or actions through the denoising process guided by the specific conditions~\cite{liao2025diffusiondrive, zhao2025diffe2e, xing2025goalflow, li2025finetuning}.

However, a shared limitation of these existing approaches is their reliance on traditional perception modules, such as obstacle detection and online HD mapping, as shown in Fig.~\ref{Fig:motivation}~(A). This reliance not only leads to a cumbersome system design but also necessitates costly perception annotations for training, with a single dataset of 1000 scenarios requiring over 7000 hours of manual labeling effort~\cite{liu2024survey}. The core principle of the end-to-end paradigm, which involves direct planning optimization via backpropagation, is intrinsically suited for training without perception annotations. While recent work has begun to explore such annotation-free paradigms~\cite{linavigation, zheng2025world4drive, lienhancing}, these methods share a critical limitation: they depend on future sensor frames and require dedicated architectures to predict latent BEV or image features for effective encoder optimization. This introduces additional complexity and potential bottlenecks. Moreover, these methods do not explicitly leverage future trajectories to construct planning-centric targets.

We propose \textit{Trajectory-oriented BEV Conditioned Diffusion~(TrajDiff)}, a framework for perception annotation-free and anchor-free generative end-to-end autonomous driving that relies solely on current sensor inputs, ego status, and future trajectories. Firstly, we formulate future trajectories as Gaussian BEV heatmap targets. As previous methods~\cite{bansal2018chauffeurnet,zhou2021exploring} verified the effectiveness of rasterized BEV representations for vehicle planning, our proposed Gaussian BEV heatmap encodes latent driving patterns, capturing the potential position and velocity characteristics. We design a lightweight Trajectory-oriented BEV Encoder to fuse raw BEV features with ego status for extracting predicted BEV heatmap and trajectory-oriented BEV feature~(TrajBEV). The former one is supervised by the Gaussian BEV heatmap, and the latter one facilitates annotation-free and anchor-free trajectory generation. This approach eliminates the need for any predefined motion priors like trajectory anchors~\cite{liao2025diffusiondrive} or goal points~\cite{xing2025goalflow}. We propose the Trajectory-oriented BEV Diffusion Transformer~(TB-DiT) module, where noise addition and denoising operations are performed directly on continuous trajectories, utilizing ego query and the TrajBEV feature as the conditions. In addition to the conventional temporal self-attention, these conditions also modulate a BEV cross-attention module, facilitating interaction between the trajectory and TrajBEV feature. Moreover, our perception annotation-free training strategy allows us to investigate the benefits of data scaling in end-to-end autonomous driving. We demonstrate that the performance of TrajDiff improves with the trajectory scaling up, even without additional scenario diversity, highlighting the robustness and scalability of our framework.

Our key contributions are summarized as follows:

\begin{itemize}
\item We propose TrajDiff, a novel end-to-end autonomous driving framework that eliminates the need for perception annotations. This is achieved by constructing Gaussian BEV heatmap targets from future trajectories, enabling self-supervised learning of effective BEV features that encode latent driving patterns.

\item We introduce a Trajectory-oriented BEV Diffusion Transformer~(TB-DiT) model that leverages TrajBEV features to denoise future trajectories in an anchor-free manner, removing reliance on handcrafted driving priors.

\item The annotation-free nature of TrajDiff uniquely enables a direct investigation into the effects of trajectory data scaling. We demonstrate that simple trajectory quantity augmentation, even without additional scenario diversity, enhances planning performance.

\item TrajDiff achieves 87.5 PDMS on the end-to-end NAVSIM benchmark, establishing a new state-of-the-art among annotation-free methods, and achieves 88.5 PDMS with simple trajectory scaling up, which is comparable to advanced perception-based methods.

\end{itemize}

\section{Related Works}
\label{sec:related}

\subsection{End-to-end Autonomous Driving}
End-to-end autonomous driving aims to directly map raw sensor inputs to planning decisions and has gained increasing research attention. UniAD~\cite{hu2023planning} proposed a purely transformer-based architecture with serialized perception-prediction-planning modules, achieving effective decision-making. VAD~\cite{jiang2023vad} explored vectorized scene representations to help autonomous vehicles narrow the trajectory search space. Transfuser~\cite{chitta2022transfuser} introduced a multi-sensor fusion framework for end-to-end driving. SparseDrive~\cite{sun2024sparsedrive} and SparseAD~\cite{sun2024sparsedrive} progressed further in the vectorized scene representation, omitting denser BEV representations. Para-Drive~\cite{weng2024drive} and DriveTransformer~\cite{jiadrivetransformer} investigated auxiliary task modules with parallelized architectures. In generative paradigms, VADv2~\cite{chen2024vadv2} introduced a large-scale planning vocabulary with scorer-based trajectory sampling. Building upon this, Hydra-MDP~\cite{li2024hydra} proposed a multi-teacher distillation scheme to inject rule-based priors into the planning decision-making. WoTE\cite{li2025wote} used a simulator to additionally supplement the closed-loop score of future scenarios for each trajectory anchor.
\begin{figure*}[t]
  \centering
  \includegraphics[width=0.97\linewidth]{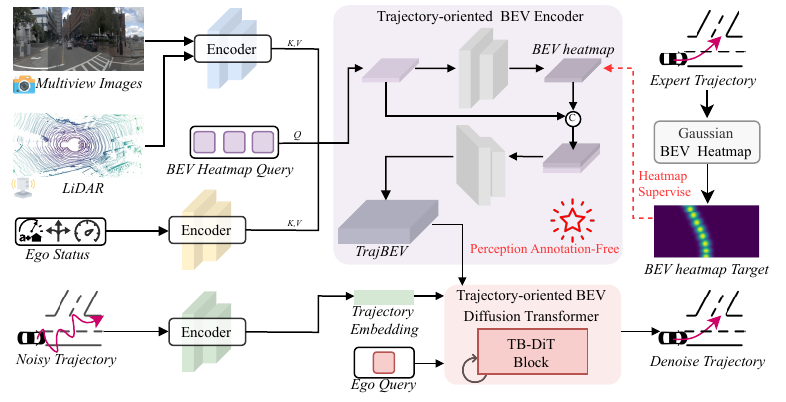}
  \caption{\textbf{Overall architecture of TrajDiff}. The Gaussian BEV heatmap is constructed from the future trajectory. The TrajBEV feature is built upon the transfuser's encoder, ego status, and heatmap query. TB-DiT accepts three key inputs: TrajBEV, ego query, and noisy trajectory. TrajDiff directly generates the predicted trajectory without prior anchor and perception annotation.}
\label{Fig:main}
\end{figure*}
\subsection{Diffusion-based End-to-end Driving}
Diffusion model, as a powerful generative framework, has achieved rapid progress in image generation and video generation. In robotic policy learning, diffusion policy demonstrates its efficacy in learning complex, high-dimensional actions for sequential decision-making~\cite{chi2023diffusionpolicy}. This success has naturally spurred its adoption in autonomous driving, where the inherent capacity for multi-modal generation is well-suited to handling uncertain traffic scenarios~\cite{zheng2025diffusionplanner, huang2024gen, yang2024diffusiones}. The effective integration of diffusion into end-to-end driving systems remains challenging. To this end, recent research has explored several key directions. DiffusionDrive~\cite{liao2025diffusiondrive} proposed a truncated diffusion policy, enabling rapid denoising over a fixed trajectory vocabulary while maintaining prediction plausibility. Meanwhile, GoalFlow~\cite{xing2025goalflow} realized a goalpoint-conditioned flow matching framework to achieve one-step trajectory generation, balancing efficiency and diversity. TrajHF~\cite{li2025finetuning} enhanced diffusion-based planning performance through a novel reinforcement learning post-training framework incorporating human feedback rewards, achieving significant performance gains. DiffE2E~\cite{zhao2025diffe2e} introduced a hybrid paradigm that synergizes generative and supervised objectives through meticulously designed loss functions, enabling an effective collaborative training strategy.

\subsection{Annotation-free End-to-end Driving}
Recent advances in end-to-end autonomous driving have shifted focus toward perception annotation-free approaches that better align with the core objective of direct planning from raw sensor inputs. Ego-MLP~\cite{li2024ego} demonstrates that ego status information alone can generate planning decisions in open-loop setting. SSR~\cite{linavigation} proves that navigation signals and a simple BEV prediction suffice for environment encoding in end-to-end planning. LAW~\cite{lienhancing} and World4Drive~\cite{zheng2025world4drive} further advanced this direction by employing future visual latent prediction and intention-aware latent prediction for self-supervised scene representation learning, eliminating perception annotation dependencies. However, current annotation-free methods still critically rely on intermediate scene prediction through BEV-based or image-based auxiliary tasks for environment understanding, necessitating both multi-frame raw sensor inputs and complex prediction modules, which fundamentally limit their scalability and efficiency. Further, these methods do not explicitly leverage future trajectories to construct planning-specific self-supervised targets.

\section{Method}
End-to-end autonomous driving aims to directly predict future trajectory, using raw sensor data and current ego status as input. The future trajectory is presented as a set of points including position and heading $X = \{(x_i, y_i, \theta_{i})\}_{i=1}^{T_f}$, where $T_f$ is the future planning horizon, and $(x_i, y_i)$ and $\theta_{i}$ are the BEV position and heading angle of ego vehicle at timestep $i$. Our work operates under a perception annotation-free setting. This constraint means that for training, the learning process relies solely on the raw sensor inputs and the ground-truth future trajectory for supervision, without access to any manual perception labels (e.g., object bounding boxes or semantic maps).

The overall pipeline of TrajDiff is illustrated in Fig.~\ref{Fig:main}. The core components of TrajDiff are a Trajectory-oriented BEV Encoder, which encodes a trajectory-oriented BEV feature~(TrajBEV) representing the potential driving pattern supervised by a Gaussian BEV heatmap target, and a Trajectory-oriented BEV Diffusion Transformer~(TB-DiT), which denoises the future planning trajectory from a noisy trajectory, controlled by the ego status and interacting with the TrajBEV. Collectively, these modules enable a simple yet effective approach to end-to-end diffusion planning within an annotation-free and anchor-free setting.

\subsection{Trajectory-oriented BEV Encoder } Leveraging the raw sensor inputs, including image and LiDAR at the current timestep, we obtain the initial BEV feature $F_{bev} \in \mathbb{R}^{H\times W \times C}$. Concurrently, for each scenario, the ego vehicle's velocity, acceleration, and navigation driving command are encoded as the ego status feature $F_{ego} \in \mathbb{R} ^ {1 \times C}$. For fair comparison, these features are obtained using the sensor backbones and transfuser encoder and ego status MLP structure consistent with those employed in \cite{chitta2022transfuser, liao2025diffusiondrive}.

Rather than optimizing BEV features through dedicated perception heads, such as those for obstacle detection or online HD mapping, we propose an alternative approach. We initialize a set of BEV heatmap queries $Q_{heatmap} \in \mathbb{R}^{N \times C}$ and employ a transformer encoder $\mathcal{G}_{H}$, where $F_{bev}$ and $F_{ego}$ are concatenated as keys and values as follow:
\begin{equation}
    \hat{Q}_{heatmap} = \mathcal{G}_{H}(Q_{heatmap}, \mathrm{Concat}(F_{bev}, F_{ego})).
\end{equation}
Computational efficiency is achieved by initializing BEV heatmap queries with spatial downsampling relative to the original BEV grid resolution. By default, we initialize the BEV queries with 4 times downsampling where $N = H \times W / 16$. Furthermore, a lightweight upsample convolutional network $\mathcal{D}_{H} $ processes these BEV heatmap queries to produce a one-channel BEV heatmap where $F_{heatmap} = \mathcal{D}_{H}(\hat{Q}_{heatmap})$ and $F_{heatmap} \in \mathbb{R} ^ {H \times W \times 1}$, which effectively represents the potential driving pattern, including position and velocity characteristics. We use the proposed Gaussian BEV heatmap target to supervise the predicted BEV heatmap in a fully self-supervised manner, and the details will be discussed in a later section. 

$F_{bev}$ and $F_{heatmap}$ are subsequently fused via a lightweight channel-fusion network $\mathcal{G}_{fuse}$ that efficiently integrates their complementary representation. The fused feature is introduced as the trajectory-oriented BEV feature~(TrajBEV), $F_{traj}$, which serves as the critical condition for our proposed diffusion transformer structure as follows:

\begin{equation}
    F_{traj} = \mathcal{G}_{fuse}(\mathrm{Concat}(F_{bev}, F_{heatmap})).
\end{equation}
\begin{figure}[t]
  \centering
  \includegraphics[width=1.0\linewidth]{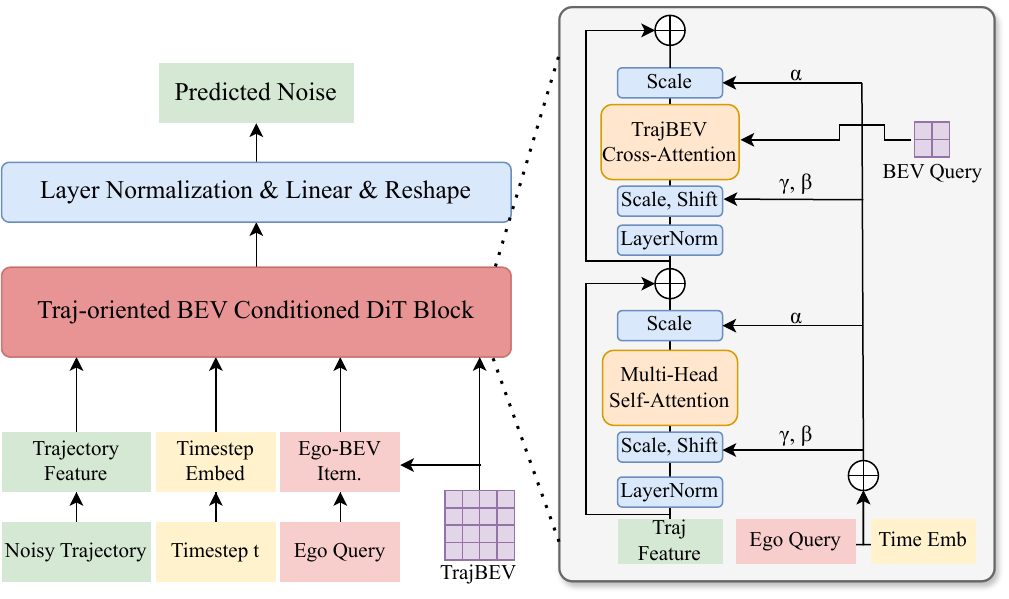}
  \caption{\textbf{Detailed architecture of TB-DiT}. TB-DiT predicts noise added on continuous trajectories by integrating the ego query and the TrajBEV feature.}
\label{Fig:dit}
\end{figure}
\subsection{Trajectory-oriented BEV DiT}
Trajectory-oriented BEV Diffusion Transformer~(TB-DiT) employs a diffusion transformer framework for anchor-free end-to-end planning. As a powerful generative approach, diffusion models effectively learn complex distributions. This is achieved through a forward process that progressively adds Gaussian noise to the input and learns a reverse process that reconstructs the clean data by iteratively denoising. In TB-DiT, we specifically perform noise addition directly on continuous trajectories. 

The forward process transforms the original future trajectories $X$ into noise over t steps:

\begin{equation}
    q(X_t | X_{0}) = \mathcal{N}(X_t; \sqrt{\bar{\alpha}_t}X_0, (1-\bar{\alpha}_t)\mathrm{I}),
\end{equation}
where $X_{0}$ is the original clean trajectory, and $X_t$ is the noised sample at noise step $t$. The hyperparameter $\bar{\alpha}_t = \prod_{s=1}^t\alpha_s = \prod_{s=1}^t(1-\beta_{s})$ and $\beta_s$ is the noise schedule. By applying the reparameterization trick, we can sample $X_t = \sqrt{\bar{\alpha}_t}X_0 + \sqrt{((1-\bar{\alpha}_t))} \epsilon_t$, where $\epsilon_t \sim \mathcal{N}(0, \mathrm{I}) $.

A dedicated noise-prediction network $\epsilon_{\theta}  $ guides the denoising procedure, where successive sampling steps generate the final clean trajectory distribution:
\begin{equation}
    X_{t-1} = \frac{1}{\sqrt{\alpha_t}}(X_t - \frac{1 - \alpha_t}{\sqrt{1 - \bar{\alpha}_t}}\epsilon_{\theta}) + \sigma_t\delta,
\label{equ:denoise}
\end{equation}
where $\delta$ is sampled from a normal Gaussian distribution and $\sigma_t$ represents the noise level. The design of an effective noise-prediction network $\epsilon_{\theta}$ constitutes the core of TB-DiT, and Fig.~\ref{Fig:dit} shows an overview of the detailed TB-DiT structure.

The inputs to TB-DiT comprise the ego query, diffusion timestep embedding, noisy trajectory, and TrajBEV feature. The ego query and timestep embedding jointly form the diffusion condition where the ego query is a learnable parameter $Q_{ego} = \mathbb{R} ^ {1 \times C}$. We introduce the Ego-BEV Interaction module $\mathcal{G}_{EB}$, composed of multiple cross-attention and MLP layers. This module enhances the interaction between the ego query and TrajBEV feature, thereby effectively fusing latent driving patterns into the ego query. The ego query is optimized as: 
\begin{equation}
    \hat{Q}_{ego} = \mathcal{G}_{EB}(Q_{ego}, F_{traj}).
\end{equation}
In TB-DiT, we employ the noise timestep embedding $F_t \in \mathbb{R}^{1 \times C}$ as conditioning, consistent with standard DiT implementation~\cite{peebles2023scalable}. Additionally, analogous to class embeddings, the ego query encoding driving pattern information $\hat{Q}_{ego}$ is incorporated as supplementary conditioning. Thus, the composite conditional embedding is formulated as follows: 
\begin{equation}
    C = F_t +  \hat{Q}_{ego}.
\end{equation}
The condition is used to predict the parameter of adaLN-Zero. TB-DiT predicts dimension-wise scaling and shift parameters from the sum of the timestep embedding and ego query. An additional input to TB-DiT is the noised future trajectory $X_t \in \mathbb{R}^{T_f \times 3}$, which serves as the primary optimization target for the diffusion denoising process. The noisy trajectory is encoded into latent trajectory features $Z_t \in \mathbb{R}^{T_f \times C}$ with a learnable trajectory encoder $\mathcal{E}_{traj}$. 

Each TB-DiT block consists of three key modules, including a temporal self-attention, a BEV cross-attention, and an MLP layer. The temporal self-attention $SA_t$ operates along the temporal dimension of the latent trajectory feature to model inter-frame interactions. A BEV cross-attention module $CA_{BEV}$ enhances trajectory-environment interaction between the latent trajectory feature and the TrajBEV feature. To reduce computational overhead while preserving critical spatial relationships, we employ a set of sparse learnable queries and a Q-former structure~\cite{li2023blip} that compresses the high-resolution BEV feature into compact query representations as $Q_{BEV}$. The processed features are then projected through a standard MLP layer $D_{traj}$ for refinement.
The computational flow within each TB-DiT block sequentially executes as follows:

\begin{equation}
    \hat{Z}_t = SA_t(Z_t),
\end{equation}
\begin{equation}
    \bar{Z}_t = CA_{BEV}(\hat{Z}_t, Q_{BEV}, Q_{BEV}),
\end{equation}
\begin{equation}
    \bar{X}_t = \mathcal{D}_{traj}(\bar{Z}_t),
\end{equation}
where $\bar{X}_t$ represents the predicted noise. Consequently, the architectural design detailed above forms the noise-prediction network $\epsilon_{\theta}$ in the diffusion model. Based on the predicted noise, the clean trajectory can be derived following Equation~\ref{equ:denoise}.


\subsection{Training Loss}

The training objective of TrajDiff integrates two key components: Gaussian BEV heatmap optimization loss and trajectory diffusion loss. Drawing upon center-based method~\cite{law2018cornernet, tian2019fcos}, we designate the ego-vehicle positions of each future timestep as the positive sample while treating all other locations as negative samples, with the penalty for negative samples increasing with their distance from the positive sample. This is because regions close to a positive sample still represent highly confident drivable areas. We construct $T_f$ Gaussian distributions in BEV space by centering each future ego-vehicle position $(x_i, y_i)$ as the mean parameters with velocity-adaptive standard deviations, effectively encoding drivable area patterns as dynamic probability fields that reflect both spatial occupancy and motion constraints as follow:
\begin{equation}
    GT_{xy} = \max_i(\exp(-\frac{(x-x_i)^2+(y-y_i)^2}{2(\Gamma v_i)^2})),
\label{groundtruth}
\end{equation}
where $v_i$ is the velocity of the ego vehicle at timestamp $i$ and $\Gamma$ is the hyperparameter controlling the speed effective radius. We utilize the Gaussian focal loss as the optimization function for the BEV heatmap prediction:
\begin{equation}
    \mathcal{L}_{bev} \!=\! -\sum_{x,y}
    \left\{\!
    \begin{array}{l@{}l}
    (1 - P_{xy})^\alpha \log (P_{xy})&\text{if } GT_{xy} \!=\! 1, \\[0.5em]
    
    P_{xy}^\alpha (1 - GT_{xy})^\gamma \log (1 - {P}_{xy})&\text{if } GT_{xy} \!<\! 1,
    
    \end{array}
    \right.
\end{equation}
where $P_{xy}$ is the predicted heatmap $F_{heatmap}$ at location $(x,y)$ and $\alpha$ and $\gamma$ is set default as in~\cite{law2018cornernet}. 

We optimize the noise prediction network using the standard diffusion loss, which computes the MSE loss between the predicted noise and the added noise:
\begin{equation}
    \mathcal{L}_{\text{Diffusion}} = \mathbb{E} \left[ \| {\epsilon}_\theta(X_t, t, Q_{ego}, Q_{BEV}) - {\epsilon_t} \|^2 \right].
\end{equation}
The final loss function is the weighted sum of BEV loss and diffusion loss as follow:
\begin{equation}
    \mathcal{L}_{final} = \omega_1 \mathcal{L}_{bev} + \omega_2 \mathcal{L}_{\text{Diffusion}}.
\end{equation}
\section{Experiments}

\subsection{Benchmark}
We train and evaluate TrajDiff on the NAVSIM benchmark~\cite{dauner2024navsim}. NAVSIM is composed of challenging scenarios built upon the OpenScene dataset~\cite{contributors2023openscene}. The navtrain split set comprises 1192 scenarios, including 978 for training and 214 for validation, and the navtest split set comprises 136 scenarios. The dataset employs an eight-camera view system and LiDAR point clouds fused from five sensors per frame to ensure comprehensive environmental perception. NAVSIM requires predicting eight future trajectory points over a 4-second horizon. Crucially, while NAVSIM provides 2Hz object and map annotations for perception-based methods, these labels are inaccessible to annotation-free approaches like TrajDiff.

The benchmark evaluates planning performance through non-reactive simulation and closed-loop metrics. All methods are uniformly assessed using the Predictive Driving Model Score (PDMS)~\cite{dauner2024navsim} – a weighted composite of five key sub-metrics: no at-fault collisions (NC), drivable area compliance (DAC), time-to-collision (TTC), comfort (Comf.), and ego progress (EP).

\begin{table*}[!ht]
    \centering
    \setlength{\tabcolsep}{8pt}
    \resizebox{\textwidth}{!}{
    \begin{tabular}{l|c|c|c|ccccc|c}
    \toprule
    \multirow{2}{*}{\textbf{Method}} & \multirow{2}{*}{\textbf{Input}} & \textbf{Perception}  & \textbf{Anchor} &  \multirow{2}{*}{\textbf{NC} $\uparrow$} & \multirow{2}{*}{\textbf{DAC} $\uparrow$} & \multirow{2}{*}{\textbf{TTC} $\uparrow$} & \multirow{2}{*}{\textbf{Comf}. $\uparrow$} & \multirow{2}{*}{\textbf{EP} $\uparrow$} & \ \multirow{2}{*}{\textbf{PDMS} $\uparrow$}  \\
     & & \textbf{Anno}. & \textbf{Free} & & & & & \\ 
    \hline
    \hline
    UniAD~\cite{hu2023planning} & C & $\checkmark$  & $\checkmark$ & 97.8 & 91.9 & 92.9 & 100.0 & 78.8 & 83.4 \\
    PARA-Drive~\cite{weng2024drive} & C & $\checkmark$  & $\checkmark$ & {97.9} & 92.4 &  {93.0} & 99.8 & 79.3 & 84.0 \\
    VADv2~\cite{chen2024vadv2} & C & $\checkmark$  & $\times$ & 97.2 & 89.1 & 91.6 & 100.0 & 76.0 & 80.9 \\
    Hydra-MDP~\cite{li2024hydra} & C \& L & $\checkmark$  &$\times$ & {97.9} & 91.7 & 92.9 & 100.0 & 77.6 & 83.0 \\
    Transfuser~\cite{chitta2022transfuser} & C \& L & $\checkmark$  & $\checkmark$ & 97.7 & 92.8 & 92.8 & 100.0 & 79.2 & 84.0 \\
    Transfuser-DP~\cite{liao2025diffusiondrive} & C \& L & $\checkmark$  & $\checkmark$ & 97.9 & 94.2 &93.9	&100.0 & 80.2 & 85.7 \\
    DiffusionDrive~\cite{liao2025diffusiondrive} & C \& L & $\checkmark$  & $\times$ &98.2  & 96.2  & 94.7  & 100.0  & 82.2  & 88.1\\
    WoTE~\cite{li2025wote} & C \& L & $\checkmark$  & $\times$ & 98.5  & 96.8  & 94.9  & 100.0  & 81.9  & 88.3\\
    \hline
    Ego-MLP~\cite{zheng2025world4drive} & E & $\times$  & $\checkmark$ & 93.0  & 77.3  & 83.6  & 100.0  & {62.8} & {65.6} \\
    LAW~\cite{lienhancing} & C & $\times$  & $\checkmark$ & 97.2  & {93.3}  & 91.9  & 100.0  & {78.8}  & {83.8}\\    
    World4Drive~\cite{zheng2025world4drive} & C & $\times$  & $\times$ & {97.4}  & 94.3  & {92.8}  & 100.0  &  {79.9}  &  {85.1}\\
    TrajDiff & C & $\times$  & $\checkmark$ & 98.0  & 95.0 & 93.7  & 100.0  & 80.8  & 86.4\\
    TrajDiff & C\&L & $\times$  & $\checkmark$ & 98.1  & 95.9  & 94.2  & 100.0  & 81.7  & 87.5\\
    TrajDiff* & C\&L & $\times$  & $\checkmark$ & 98.1  & 97.0  & 94.3  & 100.0  & 82.7  & 88.5\\
    \bottomrule
\end{tabular}}
\caption{\textbf{End-to-end planning results on NAVSIM navtest~\cite{dauner2024navsim}.} ``Input" includes~``C": Camera, ``L": LiDAR, ``E": Ego-status only. ``Perception Anno." indicates the approaches necessitate training with perception annotations. ``Anchor Free" indicates the approaches without trajectory anchors. ``*" means TrajDiff with trajectory scaling up.}
\label{table_main}
\end{table*}

~\subsection{Implementation Details}

TrajDiff follows Transfuser's training and inference protocol, where the input consists of three forward-facing camera images cropped and concatenated to 1024×256 resolution and a rasterized LiDAR point cloud. TrajDiff is trained on the trainset of navtrain split, including 978 scenarios for 200 epochs with AdamW optimizer on 8 NVIDIA A100 GPUS and the total batch size is 512. The learning rate is $6 \times 10e^{-4}$ with a cosine learning rate scheduler. The hyperparameters $\omega_1$ and $\omega_2$ are 200 and 10 in the training loss. To validate data scaling benefits, we implement a trajectory resampling strategy with initial-point variation, and takes the full navtrain dataset, including 1192 scenarios. For inference, we employ DDIM~\cite{song2020denoising} with 20 sampling steps. Other model details and parameter settings can be found in the Appendix.

\subsection{Quantitative Comparison}
We present the planning performance comparison between TrajDiff and state-of-the-art methods on the NAVSIM navtest split in Table~\ref{table_main}. Compared to existing perception annotation-free approaches, TrajDiff demonstrates significant advantages. When fairly evaluated against self-supervised camera-only methods, including LAW and World4Drive that use future frame prediction as auxiliary tasks, TrajDiff modified with camera-only input achieves improvements of 2.6 and 1.3 PDMS, respectively, validating the effectiveness of TrajDiff's self-supervised design for planning optimization. TrajDiff demonstrates highly competitive performance, even when compared to perception annotation-dependent methods. It achieves superior PDMS margins of 4.1 over serial modules design, UniAD, and 3.5 over parallel modules design, PARA-Drive, while outperforming vocabulary-based approaches VADv2 and Hydra-MDP by 6.6 and 4.5 PDMS, respectively. TrajDiff also exceeds the Transfuser baseline by 3.5 PDMS. Under identical anchor-free diffusion settings, TrajDiff further improves upon the diffusion baseline TransfuserDP proposed in~\cite{liao2025diffusiondrive} by 1.8 PDMS. With trajectory data scaling up, TrajDiff achieves an additional 1.0 PDMS gain, matching the performance of anchor-based DiffusionDrive and WoTE while exhibiting some superior submetrics.

\subsection{Ablation Study}

\noindent\textbf{Effect of designs in TB-DiT.}~Table~\ref{tab_dit} validates the efficacy of TB-DiT with its three components: TrajBEV feature, Ego-BEV Interaction module, and BEV cross-attention in TB-DiT blocks. We first remove all three components, leading TrajDiff to be devoid of any perceptual information, resulting in severely degraded performance. This baseline confirms the critical role of perceptual feature integration in achieving plausible predictions. The removal of Gaussian heatmap supervision in the trajectory-oriented BEV encoder, which means restricting BEV feature optimization with diffusion loss alone, causes severe performance degradation to 85.3 PDMS. This result indicates that TrajBEV can learn a reasonable driving probability field, replacing traditional perception-derived BEV, enabling the perception annotation-free end-to-end framework. Subsequent incremental modules reveal that the Ego-BEV Interaction module, which enables ego-to-BEV interaction, contributes 1.4 PDMS improvement, while BEV cross-attention modeling trajectory-BEV relationships contributes 0.4 PDMS improvement. The complete integration achieves peak performance at 87.5 PDMS, with each module demonstrating non-redundant contributions to planning capability.

\begin{table}[t]
    \centering
    \setlength{\tabcolsep}{4pt}
    \resizebox{0.47\textwidth}{!}{
    \begin{tabular}{ccc|cccc|c}
    \toprule
    \multirow{2}{*}{\textbf{TrajBEV}} & \textbf{EB} &  \textbf{BEV} & \multirow{2}{*}{\textbf{NC}} & \multirow{2}{*}{\textbf{DAC}} & \multirow{2}{*}{\textbf{TTC}} & \multirow{2}{*}{\textbf{EP}} & \multirow{2}{*}{\textbf{PDMS} }  \\ 

                             & \textbf{Inter}. & \textbf{Cross}. & & & & & \\   
    \hline
    \hline
    $\times$ & $\times$ & $\times$ & 96.7 & 91.9 & 90.5 & 76.9 & 81.6 \\
    $\times$ & $\checkmark$ & $\checkmark$ & 97.7 & 94.5 & 92.9 \ & 79.9 & 85.3 \\

    $\checkmark$ & $\times$ & $\checkmark$ & 97.7 & 95.0 & 93.3 \ & 80.7 & 86.1 \\

    $\checkmark$ & $\checkmark$ & $\times$ & 97.9 & 95.8 & 93.8 \ & 81.4 & 87.1 \\

    $\checkmark$ & $\checkmark$ & $\checkmark$ & 98.1 & 95.9 & 94.2 & 81.7 & 87.5 \\

    \bottomrule
    
\end{tabular}}
\caption{\textbf{Ablation on the design of TB-DiT}. ``EB Inter." and ``BEV Cross." indicate the Ego-BEV Interaction module and the BEV cross attention module in TB-DiT.}
\label{tab_dit}
\end{table}

\begin{table}[t]
    \centering
    
    \setlength{\tabcolsep}{7pt}
    \resizebox{0.47\textwidth}{!}{
    \begin{tabular}{c@{\hspace{1mm}}c|cccc|c}
    \toprule
    \textbf{Resample} & \textbf{Extra} &  \textbf{NC}  &\textbf{DAC}  & \textbf{TTC} & \textbf{EP}  & \textbf{PDMS}  \\
    \hline
    \hline
    $\times$ & $\times $ & 98.1  & 95.9  & 94.2   & 81.7  & 87.5\\
    $\checkmark $ & $\times $ & 98.1  & 96.5      & 94.2    & 81.9  & 87.9\\
    $\times $ & $\checkmark $ & 98.2  & 96.5  & 94.2    & 81.9  & 88.1\\
    $\checkmark $ & $\checkmark $ & 98.1  & 97.0  & 94.3    & 82.7  & 88.5\\
    \bottomrule
\end{tabular}}
\caption{\textbf{Data scaling up in TrajDiff}. ``Resample" means initial-point variation resampling. ``Extra" means using the full navtrain dataset.}
\label{table_data}
\end{table}

\begin{table}[t]
    \centering
   \setlength{\tabcolsep}{10pt}
    \resizebox{0.465\textwidth}{!}{
    \begin{tabular}{c|cccc|c}
    \toprule
    \textbf{Radius} &  \textbf{NC}  &\textbf{DAC}  & \textbf{TTC}   & \textbf{EP}  & \textbf{PDMS}  \\
    \hline
    \hline
    5 &  97.9  & 95.8  & 93.7    & 81.4  & 87.2\\
    25 & 97.8  & 94.4  & 93.5    & 80.2  & 85.7\\
    Vel. & 98.1  & 95.9  & 94.2   & 81.7  & 87.5\\
    \toprule
\end{tabular}}
\caption{\textbf{The design of the Gaussian BEV heatmap target}. ``Vel." means the radius at the different timestep is velocity-aware.}
\label{table_radius}
\end{table}

\noindent\textbf{Data scaling up.} We empirically validate the performance gains from data scaling as shown in Table~\ref{table_data}. Augmenting the navtrain training set through initial-point variation yields a 0.4 PDMS improvement. Utilizing the full navtrain achieves a 0.6 PDMS enhancement. The integration of both strategies elevates TrajDiff's performance to 88.5 PDMS, which is superior to the perception-based methods. These results demonstrate the inherent potential of annotation-free end-to-end paradigms to benefit from data scaling. The consistent gains underscore that TrajDiff can extract meaningful patterns without reliance on manual labels.

\noindent\textbf{Gaussian BEV Heatmap Target.}~Another critical component in TrajDiff is the BEV heatmap supervision. As shown in Table~\ref{table_radius}, the experiment with constant-radius targets reveals that an overspecified radius leads to significant performance degradation, particularly in DAC and EP metrics. This confirms the critical role of the proposed heatmap targets in accurately modeling drivable regions. A smaller radius forces TrajBEV to focus on specific positions and still generate plausible trajectories. The velocity-aware targets yield marginal but consistent improvements in NC and TTC, demonstrating enhanced environmental perception with velocity information.

\begin{table}[t]
    \centering
   \setlength{\tabcolsep}{8pt}
    \resizebox{0.465\textwidth}{!}{
    \begin{tabular}{c|cccc|c}
    \toprule
    \textbf{Rollout} &  \textbf{NC}  &\textbf{DAC}  & \textbf{TTC}   & \textbf{EP}  & \textbf{PDMS}  \\
    \hline
    \hline
    1 &  98.1  & 97.0  & 94.3    & 82.7  & 88.5(\color{gray}{$\pm$0.000)}\\
    3 &  98.3  & 97.3  & 95.0    & 83.4  & 89.3(\color{gray}{$\pm$0.010)}\\
    5 &  98.4  & 97.6  & 95.2    & 83.8  & 89.7(\color{gray}{$\pm$0.013)}\\
    10 & 98.9 & 98.6   & 96.5    & 86.4  & 92.0(\color{gray}{$\pm$0.036)}\\
    \toprule
\end{tabular}}

\caption{\textbf{Best-of-K performance with different rollout numbers}. {\color{gray}{$\pm$}} indicates the mean value of the standard deviation.}
\label{table_multi}

\end{table}

\begin{table}[t]
    \centering
    \setlength{\tabcolsep}{5pt}
    \resizebox{0.465\textwidth}{!}{
    \begin{tabular}{c|c|ccc|c|c|c}
    \toprule
    \textbf{Method} & \textbf{Steps} &  \textbf{NC}  & \textbf{DAC} & \textbf{Comf.}   & \textbf{PDMS} & \textbf{Para.} & \textbf{FPS}  \\
    \hline
    \hline
    \multirow{3}{*}{DiffusionDrive} & 2 & 98.2  & 96.2 & 100 & 88.1 & \multirow{3}{*}{60M} &  50.7\\
    & 12 & 98.2  & 96.2 & 100 & 88.1 &      & 20.2 \\
    & 20 & 98.2  & 96.2 & 100 & 88.1 &      & 13.5 \\
    \hline
    \multirow{4}{*}{TrajDiff} & 3 & 79.6  & 64.2 & 33.8 & 38.1 & \multirow{4}{*}{65M} & 52.8\\
    & 12 & 98.1 & 97.0 & 99.3 & 88.3 &  & 25.6 \\
    & 20 & 98.1  & 97.0 &  100 & 88.5 &  &  17.1 \\
    & 40 & 98.2  & 96.9 &  100 & 88.5 &  &  9.8 \\
    \toprule
\end{tabular}}
\caption{\textbf{Inference speed comparison}. The FPS is measured on an NVIDIA L20 GPU with different denoising steps. ``Para." is the number of model parameters.}
\label{talbe:speed}
\end{table}

\begin{table}[t]
    \centering
    \setlength{\tabcolsep}{6pt}
    \resizebox{0.465\textwidth}{!}{
    \begin{tabular}{c|ccccc|c}
    \toprule
    \textbf{Traj Noise} &  \textbf{NC}  &\textbf{DAC}   & \textbf{TTC}  & \textbf{EP} & \textbf{Comf}. & \textbf{PDMS}  \\
    \hline
    \hline
    w/o noise &  98.1  & 97.0  & 94.3 & 82.7 & 100 & 88.5 \\
    $\mathcal{N}(0, 0.01)$ &  98.1 & 96.8 & 94.3 &  82.6 & 99.8 & 88.4 \\
    $\mathcal{N}(0, 1)$ &  98.1 & 96.5 & 94.1 &  82.4 & 98.3 & 88.1 \\
    \toprule
\end{tabular}}

\caption{\textbf{Robustness of TrajDiff}.~``$\mathcal{N}$" means injecting Gaussian noise into ground-truth trajectories.}
\vspace{-3mm}
\label{talbe:noise}
\end{table}

\noindent\textbf{Trajectory diversity.}~We quantitatively evaluate the diversity of TrajDiff by performing K parallel sampling rollouts. The benefit of this diversity is demonstrated by the best-of-K metrics reported in Table~\ref{table_multi}. As K increases, the PDMS score improves significantly, indicating that a larger set of sampling candidates is more likely to contain a superior solution. For instance, with ten rollouts, TrajDiff's performance reaches 92.0 PDMS. We also present the mean of the standard deviation that gradually increases with the increase of rollout number, to demonstrate the diversity of sampling results. These results confirm that TrajDiff generates a rich set of distinct yet plausible trajectories, a crucial capability for complex and uncertain driving scenarios.

\begin{figure*}[t]
  \centering
  \includegraphics[width=1.0\linewidth]{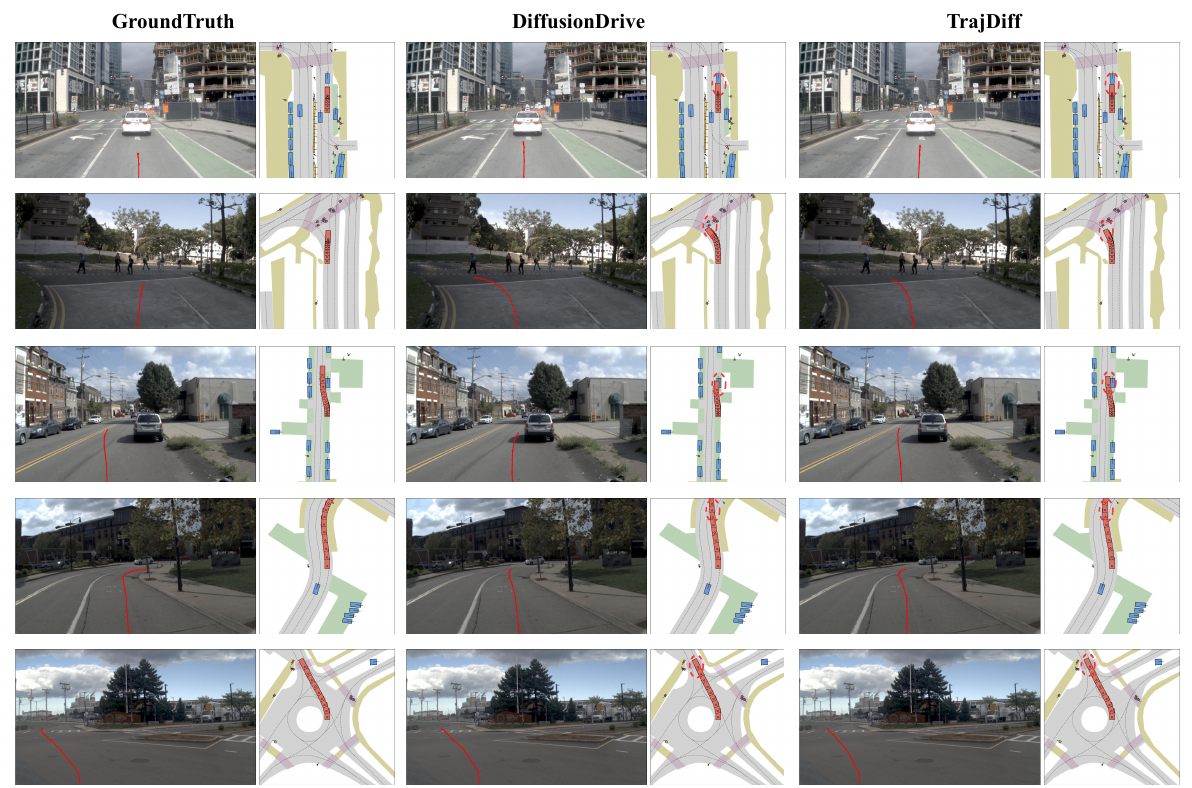}
  \caption{Qualitative comparison between the baseline methods Transfuser $\&$ DiffusionDrive and TrajDiff on the navtest split.}
\label{Fig:aa}
\end{figure*}

\noindent\textbf{Real-time feasibility analysis.}~Table~\ref{talbe:speed} details the trade-off between planning performance and inference speed with different denoise steps. We compare TrajDiff to anchor-based DiffusionDrive with a similar model size. The results show that under the same sampling step, the inference speed of TrajDiff is faster than that of DiffusionDrive. Moreover, the benefit of our anchor-free approach is a higher performance ceiling. We choose 20 steps as the default, considering a trade-off between performance and latency.

\noindent\textbf{Robustness of TrajDiff.}~We evaluate the robustness to expert trajectory noise, a crucial attribute for real-world deployment where ground-truth trajectories can be corrupted by factors like localization errors. To simulate this, we inject Gaussian noise into 10\% of the trajectory during training. For instance, the noise is sampled from a normal distribution with a standard deviation of 1 meter, representing a significant localization error and thus a challenging test of the resilience. The results in Table~\ref{talbe:noise} confirm TrajDiff's robustness against noised trajectory inputs.

\subsection{Qualitative Comparison}
Fig.~\ref{Fig:aa} presents a qualitative comparison of end-to-end planning between TrajDiff and baseline methods, including Transfuser and DiffusionDrive across three characteristic situations: deceleration, obstacle avoidance, and lane-centering. In the first two scenarios requiring stop behaviors, TrajDiff exhibits collision-aware deceleration even without obstacle annotations. The third scenario highlights its competence in collision avoidance and drivable area perception. For the final two scenarios with complex lane topologies, TrajDiff maintains plausible trajectory generation. These results validate that the TrajBEV feature enables effective environment encoding without perception annotation, and TB-DiT produces diverse and plausible motion decisions in an anchor-free manner. More visualization results can be found in the Appendix.
\section{Conclusion}
This work presents TrajDiff, a novel perception annotation-free framework for end-to-end autonomous driving. By introducing the TrajBEV feature and TB-DiT, we demonstrate that environmental understanding can be effectively learned through the Gaussian BEV heatmap, which encodes dynamic probability fields and driving patterns. The plausibility and reliability can be achieved with a dedicated diffusion transformer conditioning on TrajBEV without prior motion anchors. Moreover, data scalability is unlocked through trajectory data scaling, benefiting from the perception annotation-free setting. We will further explore end-to-end paradigms and investigate enhanced integration strategies with other functional modules.

{
    \small
    \bibliographystyle{ieeenat_fullname}
    \bibliography{main}
}


\end{document}